\documentclass[conference]{IEEEtran}
\usepackage[paper=a4paper,layout=letterpaper]{geometry}
\usepackage{url}
\usepackage{cite}
\usepackage{amsmath,amssymb,amsfonts}
\usepackage{algorithmic}
\usepackage{graphicx}
\usepackage{textcomp}
\usepackage{xcolor}
\def\BibTeX{{\rm B\kern-.05em{\sc i\kern-.025em b}\kern-.08em
    T\kern-.1667em\lower.7ex\hbox{E}\kern-.125emX}}

\usepackage{subfig}
\usepackage{times}
\usepackage{soul}
\usepackage{url}
\usepackage[hidelinks]{hyperref}
\usepackage[utf8]{inputenc}
\usepackage{amsmath}
\usepackage{booktabs}
\usepackage{algorithm}
\usepackage{algorithmic}
\usepackage{commath}
\DeclareMathOperator{\sign}{sgn}
\usepackage{mathtools}
\usepackage{amsfonts}

\ifCLASSINFOpdf
\else
\fi
\begin{document}
%
\title{Graph Spectral Feature Learning for Mixed Data of Categorical and Numerical Type}

\author{\IEEEauthorblockN{Saswata Sahoo}
\IEEEauthorblockA{Gartner\\
Gurgaon, India\\
Email: saswata.sahoo@gartner.com}
\and
\IEEEauthorblockN{Souradip Chakraborty}
\IEEEauthorblockA{Walmart Labs\\
Bangalore, India\\
Email: souradip.chakraborty@walmartlabs.com}}


%


\maketitle

\begin{abstract}
Feature learning in the presence of a mixed type of variables, numerical and categorical types, is an important issue for related modeling problems. For simple neighborhood queries under mixed data space, standard practice is to consider numerical and categorical variables separately and combining them based on some suitable distance functions.  Alternatives, such as Kernel learning or Principal Component do not explicitly consider the inter-dependence structure among the mixed type of variables.  In this work, we propose a novel strategy to explicitly model the probabilistic dependence structure among the mixed type of variables by an undirected graph. Spectral decomposition of the graph Laplacian provides the desired feature transformation. The Eigen spectrum of the transformed feature space shows  increased separability and more prominent clusterability among the observations. The main novelty of our paper lies in  capturing  interactions of the mixed feature type in an unsupervised framework using a graphical model. We numerically validate the implications of the feature learning strategy on various datasets in terms of data clusterability. 
\end{abstract}


%
\IEEEpeerreviewmaketitle

\section{Introduction}
Learning  feature transformation reducing redundant information and discovering optimal separability of the data points are important to ensure stability of the statistical models (see~\cite{noisy1} and \cite{noisy2} ). 
The data points are often mixed type consisting of numerical and categorical variables both. Feature learning can be either in a supervised framework or an unsupervised framework. In  an unsupervised framework, no ground truth is available on the class membership of the data points. The problem of feature learning in this context reduces to finding transformations of the data points  which give true separability of the data points with respect to some dissimilarity measure. In this work, we are particularly interested in an unsupervised feature learning of mixed data types.
\par
Learning features from the data can be based on linear projection on the  Principal component space introduced by \cite{pca} or based on projection on a suitable Kernel space given by \cite{kpca}. Unlike such global approaches, there is a large class of methods involving locality based manifold learning such as Local linear embedding \cite{lle} or Isometric feature mapping \cite{isomap}. Instead of the values of the data points, using the dissimilarities of the data points, Multidimensional scaling can give feature representation of the data (\cite{mds}). Singular value decomposition based approach is  used to get low rank representation of the data matrix which also gives feature representation of the data points preserving the pairwise dissimilarities (\cite{svd}). There is another class of feature learning methods which starts by constructing a suitable affinity matrix among the variables summarizing the probabilistic dependence structure among them using an appropriate graphical model. The feature representation is given by embedding the data points on the eigen space of the graph Laplacian Matrix (\cite{graph}). There are numerous recent developments in the domain of Deep Learning feature representation with high level of abstraction of the input data points giving  generalizations at different levels (~\cite{dl}).
\par
In the context of mixed space, not many techniques on feature learning are available in the literature which explicitly incorporates the interactions and dependence structure among the categorical and numerical variables. Matrix factorisation based approaches represent the mixed type data points in terms of a number of latent features, implicitly incorporating the interactions. However, such representations do not model the probabilistic dependence among the mixed type of variables as observed over the sample data points and are often difficult to interpret. In this work, we propose a method of modeling the joint dependence structure of the mixed type of variables by a suitable graphical model. The data points on the mixed variables are assumed to be independent random samples from an underlying joint probability distribution. An association between the categorical and numerical variables is defined by projecting the numerical and categorical variables on a common space. The graph structure on the projected space of the variables are estimated by maximizing the likelihood function. The estimated weighted edges essentially encode the conditional dependence structure among the variables. The desired feature transformation reduces to embedding the mixed data points on the eigen space of the Graph Laplacian. 
\par
Since the proposed feature transformation is under an unsupervised framework we also assess the effect of the transformation on data separability and cluster-ability. Previous research works on clustering mixed type of data are very limited in numbers. \cite{sbac} proposed similarity based agglomerative clustering which performs well for mixed type of data. Several attempts have been made to extend K-means clustering for categorical data by considering the cluster mode or most centrally located observation of the cluster, termed as medoid ( see \cite{pam}, \cite{proto}). These strategies largely consider numerical and categorical variables separately and eventually combining them by suitable dissimilarity measures. Several authors preferred to consider two separate clustering process for numerical and categorical data and finally combining the clustering results to obtain better partition by some ensemble method (\cite{ensemble}). The proposed feature transformation in this work is shown to increase the data separability in the transformed feature space by increasing the eigen value separation as compared to the  linear projection on the naive principal component space. The general effect of the feature transformation is shown to be positively influencing the cluster-ability of the data. The comparison of the results obtained with K-means clustering on the transformed feature space with that of the existing clustering techniques for mixed type of data is quite promising. We exhibit the results on various publicly available datasets.
\par
The rest of the paper is organized as follows. In {\bf section II} we describe the general framework and formulate the problem. {\bf Section III} is dedicated to describing the proposed methodology of the feature transformation. Finally, in {\bf section II} we describe the effect of the feature transformation on data separability and performance of the clustering.
\begin{figure}[h]

    \centering
    \includegraphics[width=0.35\textwidth]{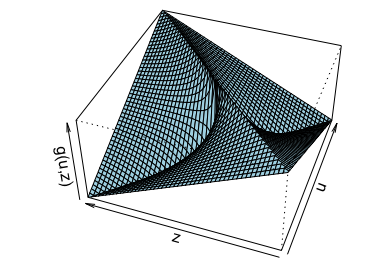}
    \caption{Similarity Function $g(u,z)$ for $u,z\in[-2,2].$}
    \label{fig:mesh1}
    \vspace{-0.2in}

\end{figure}

\section{General Framework}
Let us consider the mixed type of random variables denoted by the random vector $X=(X^{(num)'},X^{(cat)'} )'$ of dimension $p$. The subset of the variables $X^{(num)}$ is a $p_1$ dimensional random vector indicating $p_1$ numerical variables and the subset of variables $X^{(cat)}$ is a $p_2$ dimensional random vector indicating the categorical variables. The numerical random vector assumes value in the  $p_1$ dimensional Euclidean space denoted by $\mathbb{R}^{p_1}$. We assume that each of the categorical variables assumes boolean values, +1 and -1, respectively for presence or absence of the corresponding category. Hence the categorical random vector $X^{(cat)}$ has range $\{-1,1\}^{p_2}$. There are $n$ independent $p$ dimensional observations on the random vector $X$ denoted by $x_{(i)}=(x_{(i)1},x_{(i)1}, \ldots, x_{(i)p})'$, $i=1,2,\ldots,n$. The complete $n\times p$ dimensional data matrix of the $n$ observations is denoted by $\mathcal{D}=[x_{(1)}, x_{(2)},\ldots, x_{(n)}]^T$, the rows being the independent observations on the random vector $X$. The data matrix $\mathcal{D}$ is partitioned into two submatrices as $\mathcal{D}=[\mathcal{D}^{(num)},\mathcal{D}^{(cat)}]$. The submatrix $\mathcal{D}^{(num)}$ is of dimension $n\times p_1$, consisting of the $n$ rows of the numerical observations on the $p_1$ dimensional numerical random vector $X^{(num)}$ and the submatrix $\mathcal{D}^{(cat)}$ is of dimension $n\times p_2$, consisting of the $n$ rows of the categorical observations on the $p_2$ dimensional categorical random vector $X^{(cat)}.$ The main problem is to come up with a feature transformation $\phi(.)$ which transforms the $p$ dimensional mixed observation $x_{(i)}$ to $\phi(x_{(i)})$ retaining majority of information available in the sample data matrix $\mathcal{D}$.

\begin{figure}[h]
    \centering
    \includegraphics[width=0.5\textwidth]{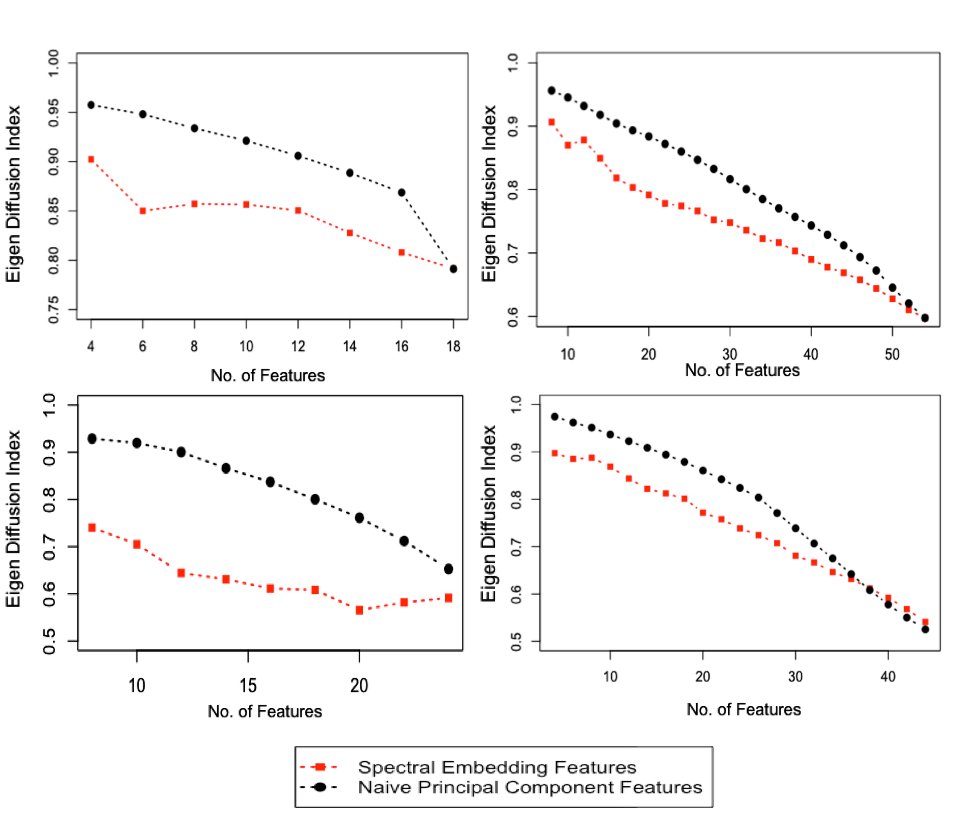}
    \caption{Plot of Eigen Diffusion index ($\alpha$) for varying choice of number of features for the proposed spectral embedding features (in red) and naive principal component features (in black) for the 4 datasets:
Clockwise starting from top left: StatLog Heart disease dataset, Teaching assistant evaluation dataset, Credit approval dataset, Adult Data Set }
    \label{eigen}
\end{figure}

\section{Methodology}
The core of the proposed methodology is to learn the probabilistic dependence structure of  the components of the mixed random vector $X$. We assume the components of $X$ as vertices of an undirected graph. The edges of the graph indicate the conditional dependence structure among the random variables corresponding to the vertices of the graph. To model the conditional dependence structure, we map the edges of the graph to the real line by a suitable function. Note that, the observable random vector consists of both numerical and categorical random variables. Mapping the edge between vertices corresponding to  a pair of numerical random variables or a pair of categorical random variables to the real line is relatively direct. However, mapping the edges passing between two vertices of a numerical and a categorical variable is not so obvious. A suitable projection strategy is introduced which maps the categorical observations on a continuous dense space. Considering the projected dense space of the categorical observations along with the actual mixed observation space $\mathbb{R}^{p_1}\times \{-1,1\}^{p_2}$, we map the edges of the graph on the real line using a suitably defined edge potential function. The graphical model parameters are estimated maximizing the Pseudo likelihood function based on the observed sample data points given in $\mathcal D.$ Considering the observations on mixed random variables as a signal over the vertices, the desired feature transformation is defined as the Fourier Transformation on the estimated Graph.
\subsection{Graphical Model}
Let us consider an undirected graph $\mathcal{G}=(\mathcal{V},\mathcal{E})$ with vertices $\mathcal{V}=\{1,2,\ldots, p\}$ and edge set $\mathcal{E}$. Each random variable of the $p$ dimensional mixed random vector $X$ is essentially corresponding to one of the vertices in $\mathcal{V}$. The joint distribution of the random vector $X$ is assumed to factorize as $f(x)\propto exp[\sum_{(s,t)\in \mathcal{E}}\psi(x_s,x_t)],$ where $x=(x_1,x_2,\ldots, x_p)'$ is an observation on $X$. 
\begin{table}
\centering
\begin{tabular}{cccc}
\hline
Method  &No. of& Rand& Total \\
 &Clusters&Index($R$) &Entropy($E$) \\
\hline
&2&{ 0.701}&{ 0.935}\\
{{\it K-Means-SE}}       &5 & { 0.587}  &{ 2.255}    \\
&10&{ 0.555}&5.011\\
\hline
&2&0.704&0.852\\
{\it K-Means-PC}       &5&0.582  &2.431    \\
&10&0.541&4.581\\
\hline
&2&0.653&1.061\\
{\it K-Proto}   &5   & 0.582 &2.261     \\
&10&0.554&4.861\\
\hline
&2&0.654&1.001\\
{\it K-Med}   &5 &0.568  &2.301   \\
&10&0.547&4.641\\
\hline
\end{tabular}
\caption{Clustering  quality with {\it K-Means-SE} in comparison to the alternatives on StatLog Heart disease dataset. }
\label{heart}
\vspace{-.2in}
\end{table}
For each edge $(s,t)\in \mathcal{E}$, the function $\psi(x_s,x_t)$ is a mapping of the edge to the real line. We consider a special case: $\psi(x_s,x_t)=\theta_{st}h(x_s,x_t),$ where $\theta_{st}\in \mathbb{R}$ and $h(x_s,x_t)\in [-1,1]$ is a known function, indicating similarity between $x_s$ and $x_t$. The model parameters $\theta_{st}, s,t\in \mathcal{V} $ are considered as the weights corresponding to the edges of the graph $\mathcal{G}$. The similarity function $h(.,.)$ between the observed values of the random vector $X$ on each pair of vertices is defined suitably by separately considering pair of vertices corresponding to the numerical variables, pair of vertices corresponding to the categorical variables and pair of vertices corresponding to a pair of mixed type of variables. It is important to keep in mind that the similarity function $h(.,.)$ should be well defined in the sense that it is scaled properly and bounded in $[-1,1]$ and also, the similarities between pairs of numerical, categorical  or mixed type of variables should be directly comparable. To define the similarity function $h$ for mixed type of variables first it is important to describe the collective factorization strategy. This strategy maps observations on categorical variables to a continuous dense space, so that the mixed type of observations are directly comparable.
\subsection{Collective Factorization of Numerical and Categorical Space}
The projection of the set of categorical variables on the continuous space is achieved through collective factorization of the matrix of numerical variables $\mathcal{D}^{(num)} \in \mathbb{R}^{n\times p_1}$ and the matrix of binary categorical variables $\mathcal{D}^{(cat)} \in \mathbb{R}^{n\times p_2}$.The main idea behind the collective factorization is that both the categorical and numerical variables can be represented in a common hidden space. Such idea has been existent for latent feature learning through joint factorization in presence of observations from multiple spaces (see \cite{coldstart}). The common latent space representation of the data is done by considering a common matrix $W$ for factorization of $\mathcal{D}^{(num)} $ and $\mathcal{D}^{(cat)} $ both, by the following  optimization problem, $||.||$ being the matrix Frobenius norm :
\begin{equation}
\begin{split}
    min_{W,H_1,H_2} :  ||\mathcal{D}^{(num)} {-} WH_1|| + ||\mathcal{D}^{(cat)}  {-} WH_2|| \\
    \text{such that } W\ge 0, H_1\ge0, H_2\ge 0. 
\end{split}
\end{equation}
\begin{figure}[h]
    \centering
    \includegraphics[width=0.5\textwidth]{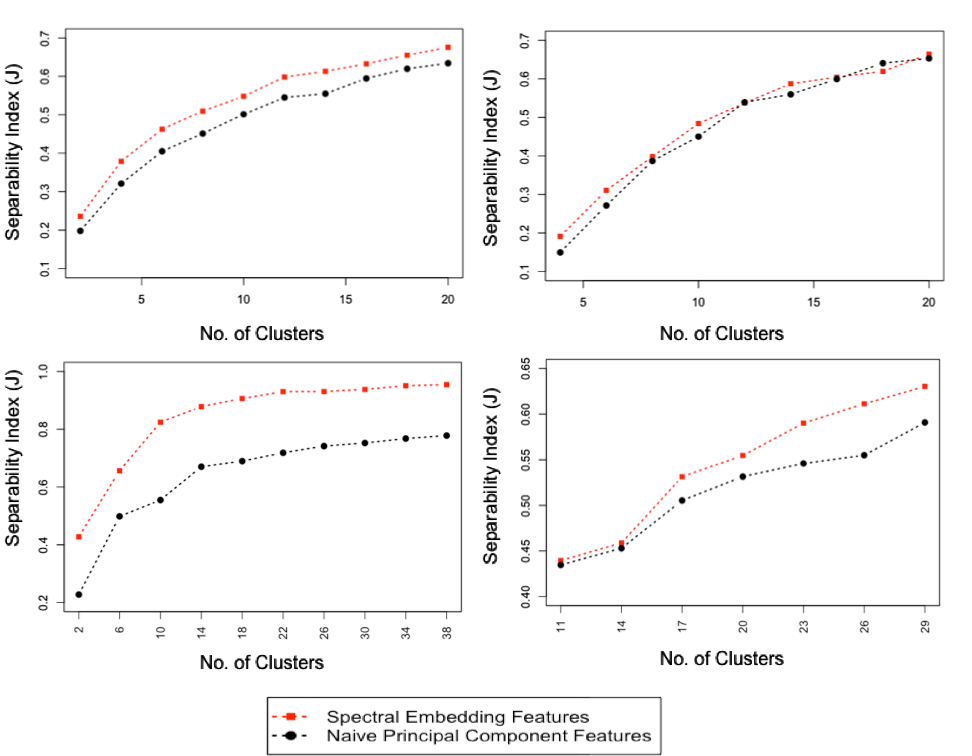}
    \caption{Plot of Cluster Separability index ($J$) under varying choice of number of clusters for the proposed spectral embedding features (in red) and naive principal component features (in black) for the 4 datasets when the number of features is fixed:
Clockwise starting from top left: StatLog Heart disease dataset, Teaching assistant evaluation dataset, Credit approval dataset, Adult Data Set }
    \label{sep}
\end{figure}
The common latent space representation is achieved by restricting $W \in R^{n\times k}$ where $k$ is the common latent dimension with $k<p_1+p_2$. In order to solve the given optimization problem in an iterative way, stochastic gradient descent has been used as given in \cite{coldstart}. 
\par
\begin{table}
\centering
\begin{tabular}{cccc}
\hline
Method  &No. of& Rand& Total \\
 &Clusters&Index($R$) &Entropy($E$) \\
\hline
&2&{ 0.486}&{ 2.119}\\
{{\it K-Means-SE}}        &5 & { 0.570}  &{ 4.805}    \\
&10&{ 0.631}&8.871\\
\hline
&2&0.490&2.122\\
{\it K-Means-PC}        &5&0.562  &5.131    \\
&10&0.626&9.281\\
\hline
&2&0.492&2.192\\
{\it K-Proto}    &5   & 0.591 &4.711     \\
&10&0.606&7.761\\
\hline
&2&0.470&2.191\\
{\it K-Med}   &5 &0.601  &5.130   \\
&10&0.647&10.146\\
\hline
\end{tabular}
\caption{Clustering  quality with {\it K-Means-SE} in comparison to the alternatives on Teaching assistant evaluation dataset. }
\label{teaching}
\vspace{-.2in}
\end{table}
Consider an observation on the mixed type of random variables denoted by $x=(x_1,x_2,\ldots, x_p)'$. The observed data point can be partitioned to $x=(x^{(num)'}, x^{(cat)'})'$ indicating the components corresponding to the numerical variables and categorical variables.
The observation on the categorical variables, denoted by $x^{(cat)}\in\{-1,1\}^{p_2}$ is given as a row of the data matrix $\mathcal{D}^{(cat)}$. The corresponding representation on the continuous space  is given by premultiplying estimated $H_2$ with the corresponding estimated row of $W$ as $\hat x^{(cat)}=\hat{w}\hat{H}_2$. The rows of estimated $W$ consists of factor loadings  of latent features of both numerical and categorical variables. Hence, to ensure comparability between the mixed spaces, the similarity between the numerical observation $x^{(num)}$ and categorical observation $x^{(cat)}$ is measured by consider the respective dense maps 
\begin{equation}\label{map}
\begin{split}
\hat x^{(num)}=\hat{w}\hat{H}_1\text{ and } \hat x^{(cat)}=\hat{w}\hat{H}_2.
\end{split}
\end{equation}

\subsection{Similarity Function }
Consider an observation on the mixed type of random variables $x=(x_1,x_2,\ldots, x_p)'$ observed over the vertices $\mathcal{V}=\{1,2,\ldots,p\}$.  We partition the vertices $\mathcal{V}$ as $\mathcal{V}=\mathcal{V}^{(num)}\cup \mathcal{V}^{(cat)}$ where $\mathcal{V}^{(num)}=\{1,2,\ldots,p_1\}$ and $\mathcal{V}^{(cat)}=\{p_1+1,p_1+2,\ldots,p=p_1+p_2\}$, respectively giving the vertices corresponding to the numerical variables and the categorical variables.  We have $x=(x^{(num)'}, x^{(cat)'})'$ and for the corresponding Collective Factorization feature map, $\hat x=(\hat x^{(num)'}, \hat x^{(cat)'})'$ as given in \eqref{map}. Note that the categorical variables only take two distinct values in $\{-1,1\}$. The observations on the numerical variables and the Collective Factorization feature map of the observations  are scaled and centred to make sure that the range of the observations are within $[-\beta, \beta]$, for some $\beta>0.$
The similarity function between  observations on a pair of variables is defined as follows. 
\begin{equation} \label{eq1}
\hspace{-0.01in}
\begin{split}
h(x_{s},x_{t}) & = x_{s} x_{t}\hspace{1.2cm} \forall s,t \in \mathcal{V}^{(cat)} \\
&=g(x_s,x_t) \hspace{0.6cm}\forall s,t \in \mathcal{V}^{(num)} \\
&=g(\hat x_s, \hat x_t)\hspace{0.6cm} \forall s \in \mathcal{V}^{(num)} ,\forall t \in \mathcal{V}^{(cat)}.
\end{split}
\end{equation}
The function $g(.,.)$ is defined as follows. For some small $\epsilon>0$,
\begin{equation} \label{eq1}
\hspace{-0.0in}
\begin{split}
 g(u,z) = \frac{\min(\abs{u},\abs{z})}{\max(\abs{u},\abs{z})} \sign(uz),\\
 & \hspace{-0.95in}if \min(\abs{u},\abs{z})>\epsilon, \\
 &\hspace{-1.45in} = \frac{\epsilon}{\max(\abs{u},\abs{z})},\\
 & \hspace{-2.2in}if \min(\abs{u},\abs{z})\le \epsilon \text{ and } \max(\abs{u},\abs{z})>\epsilon,\\
  &\hspace{-1.45in} = 1,\hspace{0.15in} if \max(\abs{u},\abs{z})\le\epsilon.
 \end{split}
\end{equation}
Note that the  similarity function $h(.,.)$ is bounded in $[-1,1]$, where -1 and +1 respectively indicate extreme dissimilarity and extreme similarity  of observations on a pair of variables. Since the defined measure of similarity is sensitive to signs of the pair of observations, the truncation at some moderately small $\epsilon>0$ is attached to ensure stable behaviour of $g(u,z)$ for $|u|$ or $|z|$ close to 0. The  function measures the similarity of observations generated on a pair of vertices in terms of the magnitude and sign both. A plot of the function $g(u,z)$ is given in Figure \ref{fig:mesh1} for $u,z\in[-2,2].$
\subsection{Model Estimation}
The joint Pseudo log-likelihood function of the model parameters $\theta=\{\theta_{st}, s,t\in V\}$ 
based on the observed data points $\mathcal{D}=[x_{(1)}, x_{(2)},\ldots, x_{(n)}]'$ is given by  
\begin{equation} \label{eq1}
\begin{split}
\log L(\theta,\mathcal{D})= \sum_{i=1}^n\mathop{\sum^{}}_{\ \forall (s,t) \in \mathcal E}\theta_{st}
h(x_{(i)s},x_{(i)t})\\ -n\log(Z(\theta)) \\
\hspace{-.3in}\text{ where }Z(\theta) =\sum\limits_{i=1}^{n}
\exp(\sum_{\mathclap{\substack{\forall (s,t) \in \mathcal E}}} \theta_{st}
h(x_{(i)s},x_{(i)t}))\\
\end{split}
\end{equation}
The parameters are estimated maximizing the Pseudo log-likelihood function based on the following update equations.
\begin{equation}
    \hat{\theta}_{st} = \theta_{st}- \eta  \frac{\partial \log L(\theta,\mathcal{D})}{\partial \theta_{st}}
\end{equation}
where the gradient direction is given as
\begin{equation}
\begin{split}
    \hspace{-1in}\frac{\partial \log L(\theta,\mathcal{D})}{\partial \theta_{st}}= \sum\limits_{i=1}^{n}
h(x_{(i)s},x_{(i)t})\\ - \frac{n}{Z(\theta)}\sum\limits_{i=1}^{N}
\exp(\sum_{\mathclap{\substack{\forall s,t \in E}}} \theta_{st}
h(x_{(i)s},x_{(i)t})) h(x_{(i)s},x_{(i)t})
\end{split}
\end{equation}
and $\eta>0$ is the learning rate. 
\subsection{Feature Transformation}
The $p$ dimensional square matrix of the model parameters are estimated maximizing the Pseudo log-likelihood function as $\hat\Theta=\big(\hat{\theta}_{st}, s,t\in\mathcal{V}\big )$. Considering the graph $\mathcal G=(\mathcal V, \mathcal E)$ where the mixed random variables in $X$ correspond to the $p$ vertices in $\mathcal V$, the conditional dependence structure among the random variables is estimated using a symmetric version of  $\hat\Theta$, given by $\tilde\Theta=\frac{1}{2}(\hat\Theta+\hat\Theta^T)=\big(\tilde \theta_{st}, s,t\in\mathcal{V}\big),$ say. This symmetric matrix is treated as the weight matrix of the edges $\mathcal E$. To ensure positive semi definiteness of the graph Laplacian of the estimated graph $\mathcal G$ in presence of possible  negative weights, \cite{negwt} suggested using 
 the absolute values, giving the weight matrix $\tilde\Theta_{abs}=\big(|\tilde \theta_{st}|, s,t\in\mathcal{V}\big).$ Considering the symmetric weight matrix $\tilde\Theta_{abs}$ of the undirected graph $\mathcal G$, we evaluate the graph Laplacian given by $\Delta=D-\tilde\Theta_{abs}$, where $D=\big(\sum_{s\ne t}|\tilde \theta_{st}|\big)$ is the degree matrix. The graph Laplacian has eigen decomposition given by $\Delta=\Phi\Lambda\Phi^T$, where $\Phi=(\phi_1,\phi_2\ldots, \phi_p)$ are the orthonormal eigen vectors and $\Lambda$ is the diagonal matrix of the eigen values. The eigen vectors essentially are the Fourier atoms on the graph and the eigen values play the role of frequencies. An observation on the mixed random vector $X$, denoted by $x$, is treated as a signal on the vertices $\mathcal V$ of the graph $\mathcal G$ and the desired feature transformation is given by $\phi(x)=\Phi^Tx.$

\begin{table}
\centering
\begin{tabular}{cccc}
\hline
Method  &No. of& Rand& Total \\
 &Clusters&Index($R$) &Entropy($E$) \\
\hline
&5&{ 0.441}&{ 2.773}\\
{{\it K-Means-SE}}        &10 & { 0.425}  &{ 5.532}    \\
&20&{0.385}&11.083\\
\hline
&5&0.436&2.761\\
{\it K-Means-PC}       &10&0.405 &5.502    \\
&20&0.379&10.911\\
\hline
&5&0.441&2.763\\
{\it K-Proto}    &10   & 0.390 &5.506     \\
&20&0.375&11.021\\
\hline
&5&0.424&2.742\\
{\it K-Med}  &10 &0.393  &5.223   \\
&20&0.373&11.011\\
\hline
\end{tabular}
\caption{Clustering  quality with {\it K-Means-SE} in comparison to the alternatives on Adult Data Set.}
\label{adult}
\vspace{-.2in}
\end{table}
\section{Numerical Investigation}
In this section we investigate the performance of the proposed feature transformation method on few datasets in terms of various metrics. The proposed spectral embedding based feature transformation denoted by {\it SE} is compared with naive principal component features, denoted by {\it PC}. To assess the quality of the proposed spectral embedding, we compare its performance with various clustering strategies suitable for mixed type of data. Along with K-means clustering based on spectral embedding features (denoted by {\it K-Means-SE}) and principal component features (denoted by {\it K-Means-PC}), we consider K-Prototype clustering and K-Medoid clustering introduced by \cite{proto} and \cite{pam} respectively. The two clustering strategies are denoted by {\it K-Proto} and {\it K-Med} respectively. K-Means clustering algorithm attempts to minimize the total sum of squares of the points with respective cluster centriods, where as, K-Modes algorithm aims to minimize the dissimilarities of the points with the respective cluster modes. The K-Prototype algorithm integrates the K-Means and K-Modes clustering methods suitable for mixed type of data. On the other hand, K-Medoid clustering algorithm targets to minimize the total dissimlarities of the points with the  most centrally located points of the respective clusters.
\subsection{Performance Metrics}
 We consider various performance metrics, separately in connection to the feature transformation itself and on its effect on the quality of clustering. For the proposed feature transformation $x\rightarrow z=\phi(x)$, which maps $p$ dimensional mixed type data point $x$ to an $l$ dimensional space using the first $l$ eigen vectors of the graph Laplacian,  denote the eigen values of the variance covariance matrix as $\lambda^{\phi}_j, j=1,2,\ldots l.$ The eigen values of the variance covariance matrix of the actual $p$ dimensional data points are given by $\lambda_j, j=1,2,\ldots p.$ 
 \subsubsection{Eigen Diffusion:}
Define a measure of diffusion of the principal component singular values associated to the $l$ dimensional feature transformation $\phi$ as  $\alpha=\frac{(\sum_{j=1}^l\lambda^{\phi}_j)^2}{l\sum_{j=1}^l\lambda^{\phi2}_j}.$ This measure is discussed by \cite{diffusion} in connection to separability of the observations and effectiveness of principal component analysis. A larger value of  $\alpha$ indicates that the observations in the transformed feature space are increasingly similar to each other making clustering a seemingly  difficult task. When the  principal component singular values of the $\phi-$ feature transformed data are arranged in decreasing order, with singular values decreasing faster, more amount of information are retained in lesser number of respective ordered principal feature dimensions. This also makes the feature transformed data non-spherical by distributing more amount of information in the directions of only few  eigen vectors.  \subsubsection{Cluster Separability:}
Under the feature transformation $\phi$, the total sum of square of the data points are given by $S^{\phi}_T=\sum_{i=1}^n(\phi(x_i)-\mu^{\phi})(\phi(x_i)-\mu^{\phi})^T,$ where the general mean of the feature transformed datapoints is given by $\mu^{\phi}=n^{-1}\sum_{i=1}^n\phi(x_i).$ For $L$ clusters of the data points the between cluster sum of squares of the feature transformed data points is given by $S^{\phi}_B=\sum_{c=1}^Ln_{c}(\mu^{\phi}_c-\mu^{\phi})(\mu^{\phi}_c-\mu^{\phi})^T,$ where the $c$th cluster has $n_c$ number of data points and the cluster mean is $\mu^{\phi}_c$ . The separability of the clusters is given by the measure $J=\frac{Tr(S^{\phi}_B)}{Tr(S^{\phi}_T)}$, where $Tr(.)$ is the trace operator (\cite{separability}).
 \subsubsection{Rand Index:}
  Suppose there is a ground truth information available on the class labels of the observations $x_i, i=1,2,\ldots, n$, denoted by $\mathcal{P}.$ The  information on the cluster membership of the observations are given by $\mathcal{C}.$ Consider pairs of observations after the feature transformation denoted by $z_i=\phi(x_i)$ and $z_j=\phi(x_j)$ for all $i\ne j\in\{1,2,\ldots,n\}.$ Denote the number of pairs of observations $(z_i,z_j)$, where $z_i$ and $z_j$ are in the same class of $\mathcal{P}$ and same cluster of $\mathcal{C}$ by $a$. Similarly, for pairs of observations $(z_i,z_j)$, where $z_i$ and $z_j$ are in same class of $\mathcal{P}$ and different clusters of $\mathcal{C}$,  different class of $\mathcal{P}$ and same cluster of $\mathcal{C}$ and finally at different class of $\mathcal{P}$ and different cluster of $\mathcal{C}$, the numbers are denoted by $b,c$ and $d$. Rand Index is given by $R=\frac{a+d}{a+b+c+d},$ with larger values indicating more similarity between ${\mathcal C}$ and ${\mathcal P}$ (\cite{rand}).

\begin{table}
\centering
\begin{tabular}{cccc}
\hline
Method  &No. of& Rand& Total \\
 &Clusters&Index($R$) &Entropy($E$) \\
\hline
&5&{ 0.527}&{ 2.791}\\
{{\textbf{\it K-Means-SE}}}        &10 & { 0.516}  &{ 5.311}    \\
&20&{ 0.516}&8.267\\
\hline
&5&0.545&2.691\\
{\it K-Means-PC}      &10&0.546 &3.905    \\
&20&0.526&6.651\\
\hline
&5&0.610&2.491\\
{\it K-Proto}   &10   & 0.548 &4.732     \\
&20&0.528&7.376\\
\hline
&5&0.571&2.544\\
{\it K-Med}   &10 &0.535  &4.364   \\
&20&0.524&7.930\\
\hline
\end{tabular}
\caption{Clustering  quality with {\it K-Means-SE} in comparison to the alternatives on Credit approval dataset. }
\label{credit}
\vspace{-.2in}
\end{table}

 \subsubsection{Cluster Label Entropy:}
The $n$ observations, $x_1,x_2,\ldots, x_n$ have true class labels $y_1, y_2,\ldots, y_n$ varying over $M$ distinct class labels $\{1,2,\ldots, M\}$. We count the number of observations included in different clusters $\mathcal{C}=\{C_1,C_2,\ldots, C_L\}$ with different class labels. A desirable property of a good clustering is to ensure that the proportion of observations from different class labels included in a given cluster has low entropy. This ensures that the information on the observations from a small number of selected class labels are high in a given cluster. We define the total cluster entropy as a measure of cluster purity, given by $E=-\sum_{c=1}^L\sum_{j=1}^Mp_{cj}log(p_{cj})$ where $p_{cj}$ is the proportion of total number of observations of the $c$th cluster with true class label $j$. A low value of entropy indicates more cluster purity. 
 \subsection{Results}
We consider 4 publicly available datasets to investigate the performance of the proposed feature transformation in terms of its impact on the eigen spectrum and data clusterability. We consider 
StatLog Heart disease dataset, Teaching assistant evaluation dataset, Adult dataset and  
Credit approval dataset (see (\cite{datasets})). Based on the proposed strategy, the feature transformation is carried out on the mixed observations and the observations are clustered by K-means clustering method using the transformed features. We compare the separability of  the covariance matrix eigen values of the transformed features with the eigen values of the covariance matrix  of the actual observations. The eigen diffusion measure ($\alpha$) is plotted in Figure \ref{eigen} for multiple datasets. A strict reduction of eigen diffusion measure is visible under the proposed strategy from the naive PCA, for varying choices of number of feature dimensions. This clearly indicates larger eigen separability which also favours data separability and clusterability in the transformed feature space. In Figure \ref{sep} we plot the Cluster Separability index $(J)$ for the proposed feature transformed data (under K-Means-SE) and naive principal component features of the data (under K-Means-PC) for varying choices of number of clusters. Expectantly, for all the datasets under investigation, a strict increase in cluster separability is visible under the proposed feature transformation. We compare the performance of K-Means clustering using the proposed feature transformed data (K-Means-SE) with naive Principal component K-Means clustering (K-Means-PC), K-Medoid clustering and K-Prototype clustering for varying choices of number of clusters. Table \ref{heart}-\ref{credit}  show the relevant results on Rand index ($R$) and Total Cluster Entropy ($E$) for the 4 datasets under investigation. Increase in Rand index and decrease in Total Cluster Entropy both favour cluster purity. In case of all the data sets, the results in terms of these two metrics for the K-Means-SE are comparable with the other competitors and sometimes are even better. In terms of the Rand index ($R$) for the first three datasets (see Table \ref{heart}-\ref{adult} ) there is a visible advantage of the proposed strategy over the competitors unless few exceptions. Otherwise, the results are in general quite comparable with the state of the art mixed data clustering strategies which in general opens a very promising direction of further investigations.

\bibliographystyle{IEEEtran}
\bibliography{ijcai19}
\end{document}